\documentclass[preprint,5p,times,twocolumn]{elsarticle}

\usepackage{amssymb}

\usepackage{booktabs}
\usepackage{tabularx}
\usepackage{hyperref}
\usepackage{amsmath, amssymb}
\usepackage{graphicx}
\usepackage{array}  
\usepackage{booktabs}  
\usepackage{multirow}  
\usepackage{amsmath}  

\biboptions{numbers,sort&compress}

\journal{Computers in Biology and Medicine}

\begin{document}

\begin{frontmatter}



\title{Modeling Long Sequences in Bladder Cancer Recurrence: A Comparative Evaluation of LSTM,Transformer,and Mamba}


\author[1]{Run-Quan Zhang}
\ead{107552203540@stu.xju.edu.cn}

\author[2]{Jia-Wen Jiang}
\ead{107552303499@stu.xju.edu.cn}

\author[3]{Xiao-Ping Shi \corref{cor1}}
\ead{shixiaoping@xju.edu.cn}

\address[1]{Department of Mathematics and Systems Science,Xinjiang University,Urumqi,Xinjiang,China}
\address[2]{Department of Mathematics and Systems Science,Xinjiang University,Urumqi,Xinjiang,China}
\address[3]{Department of Mathematics and Systems Science,Xinjiang University,Urumqi,Xinjiang,China}

\cortext[cor1]{Corresponding author}


\begin{abstract}
\textbf{Objective:} Traditional survival analysis methods often struggle with complex time-dependent data,failing to capture and interpret dynamic characteristics adequately.This study aims to evaluate the performance of three long-sequence models—LSTM,Transformer,and Mamba—in analyzing recurrence event data and integrating them with the Cox proportional hazards model.\\
\textbf{Method:} This study integrates the advantages of deep learning models for handling long-sequence data with the Cox proportional hazards model to enhance the performance in analyzing recurrent events with dynamic time information.Additionally,this study compares the ability of different models to extract and utilize features from time-dependent clinical recurrence data.\\
\textbf{Results:} The LSTM-Cox model outperformed both the Transformer-Cox and Mamba-Cox models in prediction accuracy and model fit,achieving a Concordance index of up to 0.90 on the test set.Significant predictors of bladder cancer recurrence,such as treatment stop time,maximum tumor size at recurrence and recurrence frequency,were identified.The LSTM-Cox model aligned well with clinical outcomes,effectively distinguishing between high-risk and low-risk patient groups.\\
\textbf{Conclusions:} This study demonstrates that the LSTM-Cox model is a robust and efficient method for recurrent data analysis and feature extraction,surpassing newer models like Transformer and Mamba.It offers a practical approach for integrating deep learning technologies into clinical risk prediction systems,thereby improving patient management and treatment outcomes.

\end{abstract}



\begin{keyword}


LSTM\sep
Transformer\sep
Mamba\sep
Recurrence event\sep
Survival analysis\sep
Variable selection

\end{keyword}

\end{frontmatter}


\section{Introduction}
In clinical medicine,recurrent events profoundly impact disease management and patient survival outcomes,necessitating precise prediction and analysis for cancer treatment,chronic disease management,and postoperative recovery \cite{sun2009regression, han2020variable}.The intrinsic temporal characteristics of these events,such as intervals and frequencies,provide crucial predictive information about disease progression and treatment efficacy \cite{chen2013estimating, amorim2015modelling, krol2016joint}.Classical methods like the Cox proportional hazards model \cite{cox1972regression}and the Andersen-Gill model \cite{andersen1982cox}have achieved significant accomplishments in survival and recurrent event data analysis.However,the time dependency and high-dimensional features of clinical recurrence data challenge the effectiveness of existing models.To address these issues effectively,it is imperative to develop risk prediction models that can more accurately predict and analyze the dynamic information of disease recurrence.\par

Over the past decade,deep learning techniques have revolutionized the field of survival analysis.Models such as DeepSurv \cite{katzman2018deepsurv}and RNN-SURV \cite{giunchiglia2018rnn}have enhanced neural networks'capabilities in handling nonlinear associations and censored data,thereby improving the accuracy and efficiency of survival analysis.Additionally,researchers have proposed a new deep learning framework \cite{gupta2019cresa}for survival analysis of recurrent events with multiple competing risks.This framework employs LSTM networks to improve the prediction of event times and their causes.\par

In recent years,the introduction of models such as Transformer \cite{vaswani2017attention}and Mamba \cite{gu2023mamba}has further pushed the boundaries of data processing,especially in handling long sequence data and capturing complex temporal dependencies.Transformers,with their multi-head attention mechanisms,have shown remarkable performance in various sequence modeling tasks \cite{zhou2021informer},while Mamba's selective state space approach provides efficient handling of long sequence data with linear time complexity \cite{zhu2024vision}.These models present new opportunities for improving predictive accuracy and model robustness in clinical applications.\par
\begin{table*}[h]
\centering
\begin{tabularx}{\textwidth}{lX}
\toprule
\textbf{Statement of Significance}&\\
\midrule
\textbf{Problem} & Traditional methods struggle with predicting recurrent clinical events due to complex,time-dependent data in diseases like bladder cancer.The emergence of deep learning models for long sequence data necessitates evaluating their clinical applicability and performance.\\
\textbf{What is Already Known} & Recent models like Transformer and Mamba show promise for long sequence data,but it is crucial to compare them with established models like LSTM,known for its effectiveness in longitudinal data analysis.\\
\textbf{What This Paper Adds} & This study compares LSTM,Transformer,and Mamba models for recurrence event data.The LSTM-Cox model shows superior accuracy and model fit,effectively identifying significant clinical risk factors.\\
\bottomrule    
\end{tabularx}
\caption{Statement of Significance}
\end{table*}
This study focuses on comparing the performance of LSTM,Transformer,and Mamba models in analyzing bladder cancer recurrence data,a domain characterized by high recurrence rates and rapid disease progression.Previous studies have highlighted factors such as tumor count and static tumor shape as crucial predictors of early recurrence \cite{jeong2022clinical}.Emerging studies on the prognostic utility of nuclear features extracted through machine learning also reveal the potential of advanced computational techniques in predicting non-muscle invasive bladder cancer recurrence \cite{teoh2022recurrence}.However,despite the emergence of new models,LSTM model continues to perform exceptionally well in many applications,demonstrating its enduring relevance and applicability.\par

By systematically comparing these advanced models,this study found that the LSTM model combined with the Cox model performed best in handling time-sensitive clinical recurrence data.The LSTM model outperformed the Transformer and Mamba models in terms of prediction accuracy and model fit,and its results were consistent with actual clinical trial outcomes,identifying key risk factors such as treatment stop time,maximum tumor size at recurrence,and recurrence frequency as significant predictors of bladder cancer recurrence \cite{zachos2014tumor,mahvi2018local}.This study aims to improve risk prediction accuracy by more effectively learning temporal information.This not only enhances the understanding of the strengths and weaknesses of different models but also provides practical insights for integrating the deep learning models into clinical risk prediction systems,thereby improving patient management and treatment outcomes.

\section{Methods and Materials}
\subsection{Variable Definitions}
Before delving into the methods and materials used in this study,a summary of the key variables involved in the subsequent sections is presented in Table \ref{tab:variables_summary}.This summary aims to ensure that readers can accurately understand the data processing,model assumptions,and the terminology and symbols used throughout the analysis,thereby facilitating a better grasp of these methods and their implementation details.\par

\begin{table*}[htbp]
\centering
\caption{Summary of Key Variables Used in the Study}
\begin{tabular}{>{\centering}m{3.5cm} >{\raggedright\arraybackslash}m{2.7cm} >{\centering\arraybackslash}m{2cm} >{\raggedright\arraybackslash}m{6.8cm}}
\toprule
\textbf{Variable Type} & \textbf{Variable Name} & \textbf{Variable Symbol} & \textbf{Definition} \\
\midrule
\multirow{2}{*}{Input Variables} & Covariates Matrix & \( X \) & Matrix containing feature data for all patients, used as input to the models. \\
& Time-step Features & \( X_t \) & Features at time step \( t \), used in model calculations. \\
\multirow{2}{*}{Output Variables} & Model Output & \( \hat{f}_{\Theta}(X,t) \) & The output from the models (LSTM, Transformer, Mamba) approximating function \( f(X,t) \). \\
& Hazard Function Estimate & \( \hat{h}(t|F) \) & Estimated hazard function from the Cox model combined with the deep learning models. \\
\multirow{2}{*}{Parameter Variables} & Model Parameters & \( \Theta \) & Parameters defining the network structure and weights of the deep learning models (LSTM, Transformer, Mamba). \\
& Cox Model Coefficients & \( \beta \) & Regression coefficients of the Cox proportional hazards model. \\
Intermediate Variables & Hidden State at Time \( t \) & \( H_t \) & Hidden state of LSTM at time step \( t \), used in further calculations. \\
\multirow{3}{*}{Configuration Variables} & Number of Samples & \( N \) & Total number of patient records analyzed. \\
& Number of Features & \( F \) & Total number of features considered after data restructuring. \\
& Number of Time Steps & \( T \) & Total number of time steps in the models. \\
\bottomrule
\end{tabular}
\label{tab:variables_summary}
\end{table*}

\subsection{Dataset Description}
\begin{table}[h!]
\centering
\caption{Descriptive Statistics and Treatment Frequency of the Bladder Cancer Dataset}
\label{tab:combined_stats_vertical}
\begin{tabular}{@{}lcccc@{}}
\toprule
\textbf{Statistic} & \textbf{Min} & \textbf{Median} & \textbf{Mean} & \textbf{Max} \\ 
\midrule
Number    & 1   & 2      & 2.374 & 8   \\
Size      & 1   & 1      & 1.993 & 8   \\
Recur     & 0   & 3      & 3.595 & 9   \\
Start     & 0   & 4      & 11.2  & 52  \\
Stop      & 0   & 23     & 23.8  & 64  \\
Rtumor    & 1   & 2      & 3.07  & 8   \\
Rsize     & 1   & 1      & 1.265 & 8   \\
Enum      & 1   & 2      & 2.704 & 10  \\
\midrule
\textbf{Category} & \multicolumn{4}{c}{\textbf{Frequency}} \\
\midrule
Placebo             & \multicolumn{4}{c}{48} \\
Pyridoxine          & \multicolumn{4}{c}{32}  \\
Thiotepa            & \multicolumn{4}{c}{38}  \\
\bottomrule
\end{tabular}
\end{table}
This study utilizes a bladder cancer recurrence dataset,which is publicly available on the Kaggle platform \cite{bladder_cancer_recurrences}.The dataset provides detailed information on 118 patients,including various recurrence events occurring during treatment.It records multiple key attributes for each patient,such as the type of treatment administered (including placebo, vitamin B6, and thiotepa),the initial number of tumors,the maximum tumor size,recurrence frequency,the start and end times of each recurrence event,reasons for data censoring,tumor metrics for each recurrence interval,and event observation codes.The reasons for data censoring are classified as follows: 0 indicates no recurrence (or censored),1 indicates cancer recurrence,2 denotes death caused by bladder cancer,and 3 corresponds to death due to other or uncertain causes.\par

The dataset meticulously records time-series data related to patient recurrences,providing a rich resource for analyzing bladder cancer recurrence patterns.In this study,data from all 118 patients were selected,considering all recorded recurrence events to comprehensively evaluate the proposed combined LSTM and Cox model approach.Table \ref{tab:combined_stats_vertical} presents a descriptive statistical summary of the bladder cancer recurrence dataset,including minimum,median,mean,and maximum values.For instance,the average number of tumors (Number) is 2.374,indicating the average number of tumors at the initial diagnosis.The maximum number of recurrences (Recur) is 9,showing that the patient with the highest recurrence frequency experienced 9 recurrences.Additionally,the table lists the frequency of different treatment types.The placebo treatment (Placebo) was recorded the most frequently,with 48 instances.The choice of treatment type may impact recurrence rates and treatment efficacy,making this information potentially significant for analyzing treatment outcomes.\par

This statistical information not only reflects the clinical characteristics of the dataset but also may indicate the long-term effects of treatments and patient survival.These statistics provide a foundation for our subsequent analysis of bladder cancer recurrence patterns and the evaluation of different treatment methods' effectiveness,highlighting both the complexity of the dataset and the challenges of the research.

\subsection{Data Preprocessing}
Data preprocessing involves removing non-essential columns,such as automatically generated indexes,and converting specific categorical variables into numerical formats.For example,in the \texttt{treatment} column,treatment methods are numerically encoded as follows: placebo (1),thiotepa (2),and vitamin B6 (3).Dots in the \texttt{rtumor} and \texttt{rsize} columns are replaced with NaNs,followed by appropriate handling of missing values.For cases with no recurrence or death,NaNs in the relevant columns are replaced with 0.Numerical features are standardized to reduce the impact of differing magnitudes.\par

To meet the models’requirements for time series data,patient records are reorganized into fixed-length sequences.This process generates a three-dimensional array \(D \in \mathbb{R}^{N \times T \times F}\).The sequence length is set to 3,and selected features include treatment type,number of tumors,size,and recurrence counts.These sequences are then converted into NumPy arrays and saved as \texttt{.npy} files for subsequent analysis.

\subsection{Model Design and Feature Examination}
\begin{figure*}[h]
    \centering
    \includegraphics[width=\linewidth]{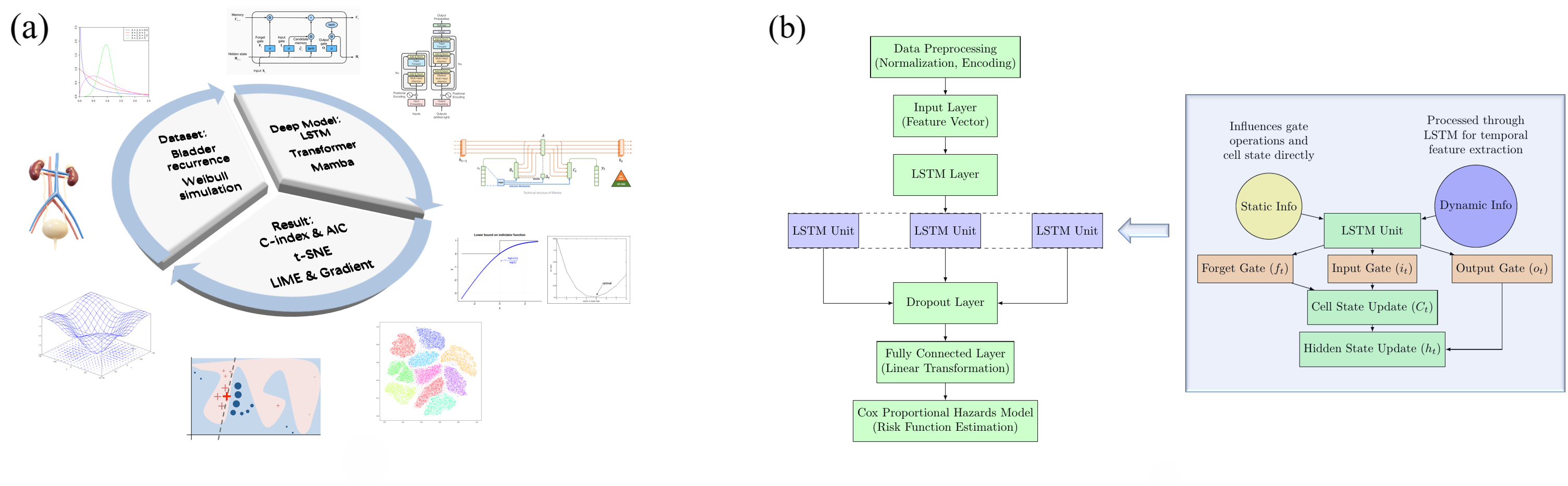}
    \caption{(a) Overall research design (b) Specific structure of the LSTM model}
    \label{fig:modeldesign}
\end{figure*}

To better illustrate the research process and model architecture, Figure \ref{fig:modeldesign} shows the overall research design in (a) and the specific structure of the LSTM model as an example in (b).

\subsubsection{LSTM Model Settings}
In this study,we constructed the LSTM model using TensorFlow.First,the input layer of the model receives the preprocessed time series data.This input can be represented as the function
\begin{equation}
f: \mathbb{R}^{T \times F} \to \mathbb{R}^M
\end{equation}
where \( M \) represents the number of output features.The model includes a 20-unit LSTM layer set to return sequences at each time point
\begin{equation}
H = [h_1, h_2, \ldots, h_T]
\end{equation}
where each \( h_t \in \mathbb{R}^U \) represents the hidden state at time step \( t \), and \( U \) is the number of LSTM units.The updates of the LSTM can be computed using the following formula
\begin{equation}
H_t = \text{LSTM}(H_{t-1}, X_t)
\end{equation}
where \( X_t \) is the input at the current time step and \( H_{t-1} \) is the hidden state of the previous time step.Additionally,to reduce overfitting,a Dropout layer is added,which randomly drops neurons using the formula
\begin{equation}
Y = \text{Dropout}(H, p)
\end{equation}
where \( p \) is set to 0.5.The final layer of the model is a fully connected layer without an activation function to suit the subsequent Cox model.\par

To quantify the discrepancy between the model predictions and the actual results,mean squared error (MSE) is used as the loss function \cite{hastie2009elements},mathematically expressed as
\begin{equation}
L(Y, \hat{Y}) = \frac{1}{N} \sum_{i=1}^{N} (Y_i - \hat{Y}_i)^2
\end{equation}
where \( N \) is the sample size.This loss function is used by the model to measure the gap between the predicted values \( \hat{Y}_i \) and the true values \( Y_i \).The model parameters are adjusted to minimize the loss function,and optimization is performed using the Adam optimizer.\par
During the training phase,the model parameters \( \theta \) are updated according to the principle of gradient descent to minimize the loss function \cite{bishop2006pattern}.The parameter update is performed as follows
\begin{equation}
\theta_{\text{new}} = \theta_{\text{old}} - \alpha \cdot \nabla_{\theta} L(Y, f(X; \theta))
\end{equation}
where \( \alpha \) is the learning rate,and \( \nabla_{\theta} \) denotes the gradient of the loss function with respect to \( \theta \).The model iterates for 100 epochs on the training dataset,with a batch size of 32,and its performance is evaluated on the validation set to monitor and prevent overfitting.After initial training,the model is evaluated on the test set,and its loss is calculated.Further performance optimization involves hyperparameter tuning,including changing the number of LSTM units,dropout ratio,and optimizer learning rate.Finally,time series features are extracted from the model to construct the Cox proportional hazards model.

\subsubsection{Transformer Model Settings}
The Transformer model employs multi-head attention mechanisms to capture complex dependencies in the time series data.The input data is standardized and represented as $X \in \mathbb{R}^{N \times T \times F}$, where $N$ is the number of samples,$T$ is the number of time steps,and $F$ is the number of features.First,the input data passes through the multi-head self-attention layer,which computes the weighted representation of the input sequence at each time step.The multi-head self-attention mechanism is calculated as follows:
\begin{equation}
\text{Attention}(Q, K, V) = \text{softmax}\left(\frac{QK^T}{\sqrt{d_k}}\right)V
\end{equation}
where $Q$,$K$,and $V$ are the query,key,and value matrices,respectively,and $d_k$ is the dimension of the key vectors.Next,the output from the multi-head self-attention layer is passed through a feedforward neural network,calculated as:
\begin{equation}
\text{FFN}(x) = \text{ReLU}(xW_1 + b_1)W_2 + b_2
\end{equation}
where $W_1$ and $W_2$ are weight matrices,and $b_1$ and $b_2$ are bias vectors.The final output is a feature vector $F_{vec}$ used for risk estimation in the Cox model.

\subsubsection{Mamba Model Settings}
The Mamba model employs a selective state space approach to handle long sequence data efficiently with linear time complexity.The input data is standardized and represented as $X \in \mathbb{R}^{N \times T \times F}$,where $N$ is the number of samples,$T$ is the number of time steps,and $F$ is the number of features.First,the input data passes through an encoding layer to generate hidden state representations $H \in \mathbb{R}^{N \times T \times D}$,where $D$ is the dimension of the hidden states.
In the selective state space layer,the state representation at each time step is computed using a state update mechanism,as follows:
\begin{equation}
H_t = A H_{t-1} + B x_t
\end{equation}
where $H_t$ is the hidden state at time step $t$,$H_{t-1}$ is the hidden state of the previous time step,$x_t$ is the input feature at time step $t$,and $A$ and $B$ are learned matrices.The final output is a feature vector $F_{vec}$ used for risk estimation in the Cox model.

\subsubsection{Construction and Training of the Cox Proportional Hazards Model}
The Cox proportional hazards model,initially proposed by Cox (1972) \cite{cox1972regression},is a well-established method in survival analysis.This model assumes that an individual's hazard function \( h(t) \) is the product of a baseline hazard function \( h_{0}(t) \) and an exponential function of the covariates \( \exp(\beta^{T}X) \),where \( \beta \) are the coefficients and \( X \) represents the covariates associated with each individual.\par
In this study,we extend the traditional Cox model to include features extracted from the LSTM,Transformer,and Mamba networks.The construction and training of the model begin with calculating the survival time \( T_i \),obtained by computing the difference between the last observation end time (stop) and the first start time (start) for each patient.Additionally,to obtain the survival status \( \delta_i \),we check the survival status at the last time point in each patient's time series data and convert it into a binary variable.
The deep learning models (LSTM, Transformer, and Mamba) process the time-dependent covariates,generating feature vectors:
\begin{equation}
 F_{vec} = f(X; \theta) 
\end{equation}
where \( \theta \) represents the parameters of the respective deep learning model,and \( X \) represents the input data.These features,along with survival times and survival statuses,are compiled into a Pandas DataFrame.\par
Subsequently,the Cox model is trained using the \texttt{CoxPHFitter} from the lifelines library,with the hazard function now defined as:
\begin{equation}
h(t | F_{vec}) = h_0(t) \exp(\beta^T F_{vec})
\end{equation}
where \( F_{vec} \) represents the features extracted by the deep learning models,with dimensions \( N \times D \).The primary objective of model training is to maximize the partial likelihood function \( L(\beta) \).After training,the model summary explains the impact of the features extracted by the deep learning models on patient risk,including the value of the concordance index,the impact of features on the hazard ratio,and the statistical significance of the coefficients.

\subsubsection{t-SNE Visualization Analysis}
To gain deeper insights into the intrinsic structure of features in the LSTM,Transformer,and Mamba models combined with the Cox model,and to distinguish risk differences among patients,this study employs the t-distributed Stochastic Neighbor Embedding (t-SNE) method for visualization analysis \cite{wattenberg2016use}.The t-SNE object implements a two-dimensional reduction of the training features,with the component parameters set to 2,a perplexity of 30,and 3000 iterations.This results in a two-dimensional feature space that reveals hidden information and clustering of samples.Using the matplotlib library,this analysis visualizes the clinical risk scores generated by the LSTM-Cox,Transformer-Cox,and Mamba-Cox models.Patients are divided into high-risk and low-risk groups based on their risk scores,with each group clearly marked in different colors to visually distinguish risk levels in the feature space.This method not only explains how the models classify patients according to risk but also visualizes the distribution of risk scores,providing concrete insights into the models'predictive capabilities and the feature space.

\subsubsection{LIME and Gradient Test Analysis}
To gain a deeper understanding of the LSTM model's contribution to predicting bladder cancer recurrence,this study employs the Local Interpretable Model-Agnostic Explanations (LIME) method.LIME provides detailed explanations of model predictions at the individual sample level,revealing the most influential features and their contributions \cite{ribeiro2016should}.Mathematically,LIME approximates the behavior of a complex model near a point using a local linear model,represented as:
\begin{equation}
\xi(x) = \arg \min_{g \in G} L(f, g, \pi_x) + \Omega(g)
\end{equation}
where \( x \) is the sample point to be explained,\( f \) is the model in this study,\( g \) is the simple model (usually a linear model),\( G \) is the class of simple models,\( \pi_x \) is a weighting function defined near the sample \( x \),\( L \) is the loss function between the predictions of the complex model and the simple model,and \( \Omega(g) \) measures the complexity of the model.\par

First,a prediction function is defined to convert the test data into a format suitable for the LSTM model and make predictions,allowing the extraction of the output for specific samples from the model.Subsequently,an explainer capable of interpreting the predictions of the complex model is created.The training data,feature names,and regression mode are input to fit the prediction requirements of this study.\par

Next,to analyze feature contributions,LIME analysis is performed for each test sample,generating corresponding feature contribution reports.These reports clarify the most important features in the model's predictions and their contributions,helping us understand the model's behavior in specific situations.Additionally,Python's Counter function is used to calculate the frequency of each feature's appearance in the LIME explanations.This not only reveals the most important features across all samples but also demonstrates their overall impact on model predictions.\par

Furthermore,to deepen the analysis of how the LSTM model processes input features at different time steps,a gradient test is conducted.This test aims to reveal the extent to which input features affect the model's hidden layer outputs,thereby providing insights into the feature sensitivity of the model at each time step \cite{sundararajan2017axiomatic}.\par

Initially,a gradient function \( G \) is defined to describe the influence of the input on the hidden state,mathematically expressed as:
\begin{equation}
G(X_t, h_{t-1}; \theta) = \frac{\partial h_t}{\partial X_t}
\end{equation}
where \( X_t \) represents the input feature at time step \( t \),\( h_{t-1} \) is the hidden state of the previous time step,\( h_t \) is the hidden state of the current time step,and \( \theta \) represents the model parameters.The training data is converted into tensors,and these gradients are calculated through the forward propagation and automatic differentiation functions of the LSTM model.\par

Subsequently,the average gradients of the input features at each time step for all samples are calculated,represented as:
\begin{equation}
\bar{G}_{X_t} = \frac{1}{N} \sum_{i=1}^{N} G(X_t^{(i)}, h_{t-1}^{(i)}; \theta)
\end{equation}
where \( N \) is the total number of samples.By analyzing these average gradients,the study identifies the features that have the greatest impact on the model's predictions.Additionally,the frequency of each feature being identified as the maximum gradient feature is counted,providing quantitative data to assess the most critical features in the model's decision-making process.\par

It is important to note that,due to the superior performance of the LSTM model in this study,both LIME and gradient test analyses were conducted solely on the LSTM model.The LSTM model's predictions were found to be more accurate and consistent compared to other models,thus warranting a more detailed interpretability analysis using these methods.

\subsection{Validation on Simulated Dataset}
To validate the LSTM-Cox,Transformer-Cox,and Mamba-Cox models,this study used the \texttt{reda} package in RStudio to create a simulated dataset following a Weibull distribution \cite{weibull1951statistical}.This simulation generated 50 patient samples,each fitted based on the Weibull proportional hazards function.By comparing event occurrence times with observation end times,the dataset had a censoring rate of approximately 40\%.The dataset included continuous and binary variables,as well as simulated survival times and event indicators,with continuous variables following a normal distribution,binary variables following a binomial distribution,and observation end times also normally distributed.Data preprocessing was performed using the \texttt{dplyr} package to optimize the dataset for survival analysis,implementing adjustments such as adding stop columns and organizing data by ID and time columns.To verify the realism of the simulated data,the Cox proportional hazards model from the \texttt{survival} package was used to assess the significance of survival variables,ensuring the dataset's suitability for survival analysis.The final step of validation involved comparing the results of the LSTM-Cox,Transformer-Cox,and Mamba-Cox models with the traditional recurrence model,evaluating model fit through AIC to demonstrate the relative robustness and accuracy of these models in survival prediction.

\subsection{Comparison of Traditional Survival and Recurrence Models}
To further evaluate the performance of the LSTM-Cox,Transformer-Cox,and Mamba-Cox models introduced in this study on the bladder cancer clinical recurrence dataset,we conducted a series of comparative analyses against several well-known and established statistical models in survival analysis.This comparison includes the standard Cox proportional hazards model,the Wei,Lin,and Weissfeld (WLW) model \cite{wei1989regression},the Andersen-Gill model,and the Prentice,Williams,and Peterson (PWP) model \cite{prentice1981regression}.Using the \texttt{survival} package in RStudio,we accurately applied these models to the same bladder cancer dataset,consistently considering variables such as treatment type,number of tumors,and size.Model evaluation employed the Akaike Information Criterion (AIC) as a metric to quantitatively assess and compare each model's predictive accuracy and goodness of fit.This approach helps the study determine the performance of the LSTM-Cox,Transformer-Cox,and Mamba-Cox models relative to traditional survival and recurrence models.

\subsection{Software and Tools}
The computations for this study were performed using Python 3.8.18,with the IDE being PyCharm 2023.2.1 (Community Edition).Deep learning tasks were conducted using PyTorch 2.0.1,and statistical analyses were performed using R 4.3.2.The Mamba model was specifically constructed using Ubuntu 22.04.3 LTS and the Linux version of PyCharm Community Edition.

\section{Results}
To evaluate the performance of the LSTM-Cox,Transformer-Cox,and Mamba-Cox models,we used a simulated dataset following a Weibull distribution.The results are summarized in Table \ref{tab:simulated_results},which compares each model's Concordance Index (C-index) and Akaike Information Criterion (AIC) values.
\begin{table}[h!]
\centering
\caption{Comparison of Model Performance on Simulated Dataset}
\label{tab:simulated_results}
\begin{tabular}{lcc}
\hline
Model                  & Concordance Index & AIC     \\ \hline
LSTM-Cox               & 0.75              & 144.98  \\
Transformer-Cox        & 0.69              & 150.05  \\
Mamba-Cox              & 0.59              & 156.34  \\ \hline
\end{tabular}
\end{table}
The LSTM-Cox model achieved the highest C-index and lowest AIC on the simulated dataset,indicating its superior predictive accuracy and model fit compared to the Transformer-Cox and Mamba-Cox models.Additionally,the LSTM model was able to capture more significant mixed variables in the simulated dataset,whereas the Transformer model only captured one significant mixed variable,and the Mamba model could not capture any significant mixed variables.\par

The performance of the LSTM-Cox,Transformer-Cox,and Mamba-Cox models was also evaluated on the real bladder cancer recurrence dataset.Table \ref{tab:real_results} shows the C-index and AIC values for these models,as well as traditional models such as Andersen-Gill (AG),Prentice-Williams-Peterson (PWP),and Wei-Lin-Weissfeld (WLW).\par
\begin{table}[h!]
\centering
\caption{Comparison of Model Performance on Bladder Cancer Recurrence Dataset}
\label{tab:real_results}
\begin{tabular}{lcc}
\hline
Model                  & Concordance Index & AIC     \\ \hline
LSTM-Cox               & 0.90              & 221.63  \\
Transformer-Cox        & 0.92              & 204.15  \\
Mamba-Cox              & 0.63              & 299.12  \\
Andersen-Gill          & 0.78              & 212.94  \\
PWP                    & 0.74              & -       \\ 
WLW                    & 0.78              & 212.94  \\
Cox                    & 0.71              & $>$500.00 \\ \hline
\end{tabular}
\end{table}
Although the Transformer-Cox model slightly outperformed the LSTM-Cox model in terms of C-index and AIC,the LSTM-Cox model still showed very close performance and was significantly better than the Mamba-Cox model in many aspects.This demonstrates the robustness and lasting applicability of the LSTM model on the real dataset.Furthermore,the LSTM model was able to capture more significant mixed variables in the real dataset,while the Transformer model also captured more significant mixed variables,and the Mamba model was unable to capture any significant mixed variables.\par

\begin{figure*}[h!]
\centering
\includegraphics[width=\linewidth]{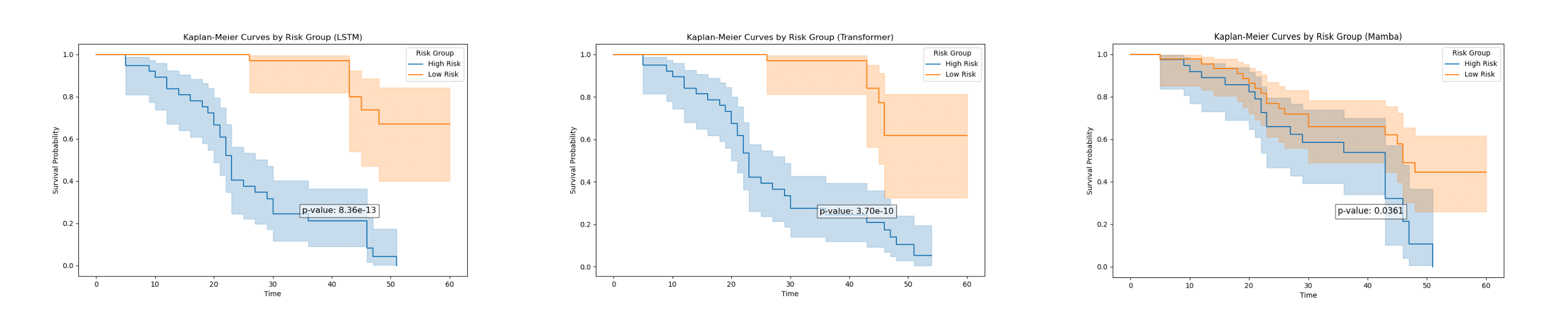}
\caption{Kaplan-Meier curves for high-risk and low-risk groups predicted by the LSTM-Cox,Transformer-Cox,and Mamba-Cox models}
\label{fig:km_curves}
\end{figure*}

\begin{figure*}[h!]
\centering
\includegraphics[width=\linewidth]{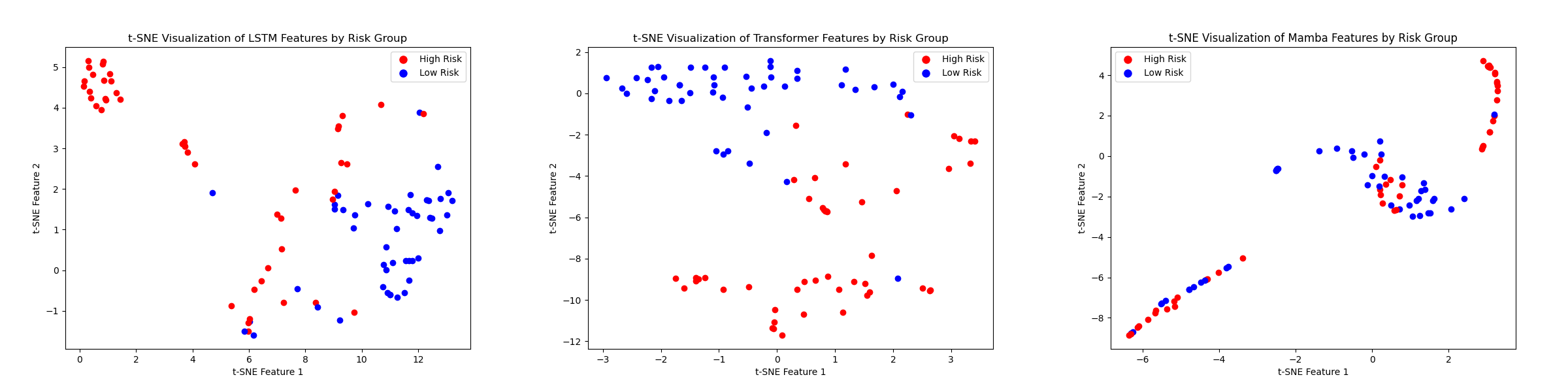}
\caption{t-SNE visualizations of patient risk profiles based on features extracted by the LSTM-Cox,Transformer-Cox,and Mamba-Cox models}
\label{fig:t-SNE_all}
\end{figure*}

To further illustrate the models'performance,Kaplan-Meier (KM) curves and t-SNE visualizations were generated.Figure \ref{fig:km_curves} shows the KM curves for high-risk and low-risk groups predicted by the LSTM-Cox,Transformer-Cox,and Mamba-Cox models.Figure \ref{fig:t-SNE_all} displays the t-SNE visualizations of patient risk profiles based on features extracted by each model.\par

From Figure \ref{fig:km_curves},it can be seen that both the LSTM-Cox and Transformer-Cox models showed significant differences in survival probability between high-risk and low-risk groups.The p-value between the high-risk and low-risk groups for the LSTM-Cox model was 8.36e-13,and for the Transformer-Cox model,it was 3.70e-10,indicating strong statistical significance in distinguishing high and low-risk groups.In contrast,the Mamba-Cox model had a p-value of 0.0361,which,although significant,showed less distinction capability than the other two models.\par

From Figure \ref{fig:t-SNE_all},it is evident that the LSTM-Cox model showed clear separation between high-risk and low-risk patients,with red dots representing high-risk patients and blue dots representing low-risk patients.The Transformer-Cox model also exhibited significant separation in distinguishing high-risk and low-risk patients,whereas the Mamba-Cox model's separation was less effective,showing considerable overlap.The performance of the LSTM-Cox model in the feature space further proves its robustness and applicability in practical applications.\par

\begin{figure*}[h!]
\centering
\includegraphics[width=\linewidth]{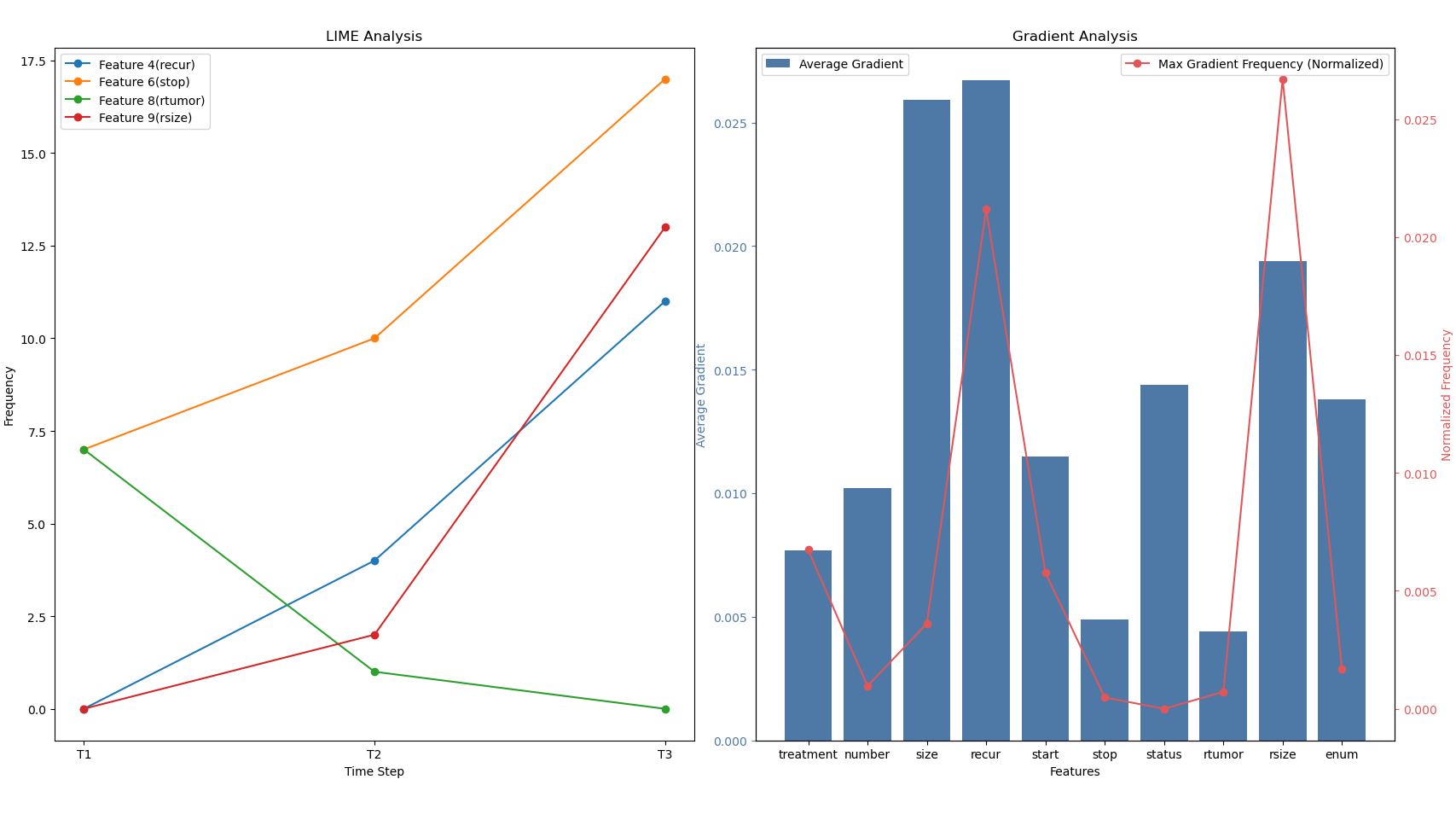}
\caption{LIME and Gradient analysis displaying the frequency and impact of features on LSTM-Cox model predictions}
\label{fig:lime_gradient}
\end{figure*}

To explore the relationship between the clinical mixed features extracted by the LSTM-Cox model and the original dataset features,the LIME results significantly revealed the contribution of the original dataset features to the LSTM-Cox model's predictions (Figure \ref{fig:lime_gradient}).An in-depth analysis of test samples highlighted the importance of features 4 (recur),6 (stop),and 9 (rsize) extracted by the neural network in the model's predictions,underscoring their critical role in risk prediction.\par

Additionally,based on the recurrence count information of samples,this study tabulated the time step data for each sample up to a maximum of the first three recurrences.By analyzing the frequency of occurrence of all features in the LIME explanations,a comprehensive understanding of the impact of each clinical feature of bladder cancer on overall risk prediction was provided.In this context,feature 6 was identified as the most important,appearing 51 times,followed by feature 9 (23 times) and feature 4 (22 times).Furthermore,feature 8 (rtumor) appeared 13 times.The LIME test results emphasized the significant impact of specific features (especially 4, 6, 8, and 9) in the LSTM-Cox model,deepening our understanding of the model's decision-making process in handling complex time series data.\par

Gradient tests conducted on the LSTM model revealed the influence of different features on model output.These tests showed that the \texttt{recur} and \texttt{rsize} variables significantly impact the model's output,underscoring their importance.Statistical analysis of the frequency of gradient features further confirmed that \texttt{rsize} and \texttt{recur} are the primary factors influencing model predictions,highlighting the critical role of these variables in determining clinical recurrence and survival outcomes in bladder cancer patients.\par

Overall,the LSTM-Cox model demonstrated superior performance and interpretability,making it a robust tool for predicting bladder cancer recurrence and showing its lasting applicability under various conditions,not inferior to the Transformer and Mamba models.

\section{Discussion}
This study highlights the comparative performance of LSTM,Transformer,and Mamba models on both simulated and real bladder cancer clinical recurrence datasets.Despite the emergence of new models like Transformer and Mamba,the LSTM model,an established method,demonstrated superior performance.From a comprehensive comparison of both datasets,the LSTM-Cox model was slightly better than the Transformer-Cox model and significantly better than the Mamba-Cox model.This suggests that traditional models like LSTM,when appropriately combined with Cox models,can still outperform newer models in certain contexts \cite{wang2021maintenance}.\par

Through rigorous evaluation,including LIME and gradient analysis,we elucidated the key mechanisms of the LSTM-Cox model in predicting clinical risk for bladder cancer recurrence.The LIME assessment particularly highlighted the significant impact of certain predictive factors such as feature 6 (treatment stop time),feature 9 (maximum tumor size at recurrence),and feature 4 (recurrence frequency).These factors are crucial in the model's risk prediction.The repeated identification of feature 6 underscores the critical role of treatment duration in predicting recurrence,while the significance of features \texttt{rsize} and \texttt{recur} reveals their important contributions to risk prediction.Notably,the significance of rsize and recur in the gradient feature frequency analysis confirmed the pivotal role of these clinical features in assessing the recurrence risk of bladder cancer patients.These findings indicate that tumor size,recurrence count,and treatment method are key prognostic indicators in the treatment and recurrence phases of bladder cancer,aligning with clinical trial results discussed in the introduction \cite{amorim2015modelling,jeong2022clinical}.Through LIME evaluation and gradient analysis,the model demonstrated its robust capability in identifying and interpreting key variables in the clinical recurrence dataset,enhancing its practicality and applicability while elevating its importance in clinical data analysis.Particularly in handling recurrence event data,this model surpasses the limitations of traditional models in a concise and effective manner,bringing innovation to this complex field.\par

Moreover,the t-SNE visualizations provided insightful differentiation between high-risk and low-risk patient groups,effectively demonstrating the practical utility of the LSTM-Cox model.The distinct separation of these groups underscores the model's robust capability in classifying patients based on risk profiles.In contrast,while the Transformer-Cox model also showed notable separation,the Mamba-Cox model exhibited less effective differentiation with considerable overlap,suggesting limitations in its feature extraction and classification abilities.This reinforces the practical relevance of the LSTM-Cox model,highlighting its robustness and applicability in real-world clinical settings.\par

Interestingly,while the Transformer-Cox model showed competitive performance,especially in capturing significant mixed variables in the real dataset,it could only capture one significant mixed variable in the simulated dataset.In contrast,the Mamba-Cox model struggled to capture significant mixed variables in both datasets.This highlights the enduring strength of the LSTM model in feature extraction and risk prediction,outperforming the newer models under various conditions.\par

The applicability of these models to broader datasets requires further validation.Future research will focus on updating the LSTM model architecture \cite{niu2018alstm,wu2019lstm},enhancing parameter validation to extend its application across various clinical recurrence datasets,and improving its interpretability.Researchers will also explore incorporating more deep learning models designed for handling long sequence data \cite{han2021transformer,gu2023mamba}.By comparing the integration differences among various models,the goal is to enable more extensive and effective selection and utilization of these methods in clinical settings.\par

In summary,the LSTM-Cox model presented in this study has been statistically validated,showing significant advantages in efficiency and usability.By deeply exploring the contributions of key clinical predictive factors,it demonstrates practicality,applicability,and innovation.These characteristics make it a powerful tool for handling clinical recurrence data,providing profound insights into integrating advanced deep learning and other computational methods with clinical decision support systems \cite{ao2023new,mou2023transformer,zou2018predicting,miao2024integrated}.The study indicates that the LSTM-Cox model,as an established method,shows lasting applicability and even outperforms newer models like Transformer and Mamba in specific contexts,proving superior to both classical recurrence models and new long-sequence deep learning models.

\section{Conclusion}
In brief,this study comprehensively compared the performance of three models for processing long sequence—LSTM,Transformer,and Mamba—in extracting features from recurrence event data and integrating them with the Cox proportional hazards model.The results demonstrate that the LSTM-Cox model outperforms the newer Transformer and Mamba models,highlighting the enduring capability and relevance of LSTM in handling longitudinal data.Despite the introduction of newer models like Transformer and Mamba,the LSTM-Cox model proved to be highly effective,showcasing its continued applicability and superior performance in many tasks.Moreover, the LSTM-Cox model provided results that align well with clinical outcomes,making it particularly effective for predicting clinical bladder cancer recurrence risk by effectively handling the time dependency in clinical recurrence data.Key risk factors such as treatment stop time,maximum tumor size at recurrence,and recurrence frequency were identified as significant predictors of bladder cancer recurrence.These findings emphasize the practical applicability of the LSTM-Cox model in clinical settings,where it can offer enhanced interpretability and usability for practitioners.

\section*{Funding}
This research did not receive any specific grant from funding agencies in the public,commercial,or not-for-profit sectors.

\section*{CRediT authorship contribution statement}
\textbf{Run-Quan Zhang:} Conceptualization,Data curation,Formal analysis,Methodology,Visualization,Writing–original draft,Writing–review \& editing.\textbf{Jia-Wen Jiang:} Writing–review \& editing.\textbf{Xiao-Ping Shi:} Conceptualization,Methodology.

\section*{Declaration of competing interest}
The authors declare that they have no known competing financial interests or personal relationships that could have appeared to influence the work reported in this paper.

\section*{Acknowledgements}
WE express our gratitude to Kaggle for hosting the dataset and to Utkarsh Singh for making the data publicly available.

\section*{Declaration of generative AI and AI-assisted technologies in the writing process}
During the preparation of this work the authors used ChatGPT-4o in order to article-polishing.After using this tool/service,the authors reviewed and edited the content as needed and take full responsibility for the content of the publication.

\section*{Appendix A. Supplementary material}
The relevant data sources are provided in the References section.


\bibliographystyle{elsarticle-num} 
\bibliography{ref}





\end{document}